\documentclass{article}

\usepackage{microtype}
\usepackage{graphicx}
\usepackage{subfigure}
\usepackage{booktabs} 
\usepackage{url}
\usepackage{enumitem}
\usepackage{amsmath}
\usepackage{amssymb}
\usepackage[export]{adjustbox} 
\usepackage{float} 
\usepackage{multicol}
\usepackage{wrapfig}

\usepackage{booktabs} 
\usepackage{pifont} 
\usepackage{hyperref}

\usepackage[accepted]{icml2021}

\icmltitlerunning{Guided Exploration with Proximal Policy Optimization using a Single Demonstration}

\begin{document}

\twocolumn[
\icmltitle{Guided Exploration with Proximal Policy Optimization using a Single Demonstration}



\begin{icmlauthorlist}
\icmlauthor{Gabriele Libardi}{upf}
\icmlauthor{Sebastian Dittert}{upf}
\icmlauthor{Gianni De Fabritiis}{upf,icrea}
\end{icmlauthorlist}

\icmlaffiliation{upf}{Computational Science Laboratory, Universitat Pompeu Fabra (UPF)
 }
\icmlaffiliation{icrea}{ICREA 
}

\icmlcorrespondingauthor{Gabriele Libardi}{gabrielelibardi@yahoo.it}
\icmlcorrespondingauthor{Gianni De Fabritiis}{gianni.defabritiis@upf.edu}

\icmlkeywords{Machine Learning, ICML, Distributed, Reinforcement Learning, PyTorch, Library}
\vskip 0.3in]


\icmlkeywords{Machine Learning, ICML}

\vskip 0.3in



\printAffiliationsAndNotice{} 

\begin{abstract}
  Solving sparse reward tasks through exploration is one of the major challenges in deep reinforcement learning, especially in three-dimensional, partially-observable environments. Critically, the algorithm proposed in this article is capable of using a single human demonstration to solve hard-exploration problems. We train an agent on a combination of demonstrations and own experience to solve  problems with variable initial conditions and we integrate it with proximal policy optimization (PPO). The agent is also able to increase its performance and to tackle harder problems by replaying its own past trajectories prioritizing them based on the obtained reward and the maximum value of the trajectory.
We finally compare variations of this algorithm to different imitation learning algorithms on a set of hard-exploration tasks in the Animal-AI Olympics environment.
To the best of our knowledge, learning a task in a three-dimensional environment with comparable difficulty has never been considered before using only one human demonstration.
\end{abstract}

\section{Introduction}

Exploration is one of the most challenging problems in reinforcement learning. Although significant progress has been made in solving this problem in recent years \citep{badia2020agent57, badia2020never, burda2018exploration, pathak2017curiosity, ecoffet2019go}, many of the solutions rely on very strong assumptions such as access to the agent position and deterministic, fully observable or low dimensional environments. For three-dimensional, partially observable stochastic environments with no access to the agent position the exploration problem still remains  unsolved.

Learning from demonstrations allows to directly bypass this problem but it only works under specific conditions, e.g. large number of demonstration trajectories or access to an expert to query for optimal actions \citep{ross2011reduction}. Furthermore, a policy learned from demonstrations in this way will only be optimal in the vicinity of the demonstrator trajectories and only for the initial conditions of such trajectories. 
Our work is at the intersection of these two classes of solutions, exploration and imitation, in that we use only one trajectory from the demonstrator per problem to solve hard-exploration tasks \footnote{A video showing the experimental results is available at \url{https://www.youtube.com/playlist?list=PLBeSdcnDP2WFQWLBrLGSkwtitneOelcm-} }$^{,}$\footnote{The code is available here: \url{https://github.com/compsciencelab/ppo_D} }.

This approach has been investigated before by \citet{paine2019making}
 (for a thorough comparison see Section \ref{chap:comparison}). We propose the first implementation based on on-policy algorithms, in particular PPO. Henceforth we refer to the version of the algorithm we put forward as PPO + Demonstrations (PPO+D). Our contributions can be summarized as follows:

\begin{enumerate}
	\item In our approach, we treat the demonstrations trajectories as if they were the result of lucky actions taken by the agent in the real environment. We can do this because in PPO the policy only specifies a distribution over the action space. We force the actions of the policy to equal the demonstration actions instead of sampling from the policy distribution and in this way we accelerate the exploration phase. We use importance sampling to account for sampling from a distribution different than the policy. The frequency with which the demonstrations are sampled depends on an adjustable hyperparameter $\rho$, as described in \citet{paine2019making}.
	
	\item Our algorithm includes the successful trajectories in the replay buffer during training and treats them as demonstrations.
	
	\item The non-successful trajectories are ranked according to their maximum estimated value and are stored in a different replay buffer.
	
	\item We mitigate the effect of catastrophic forgetting by using the maximum reward and the maximum estimated value of the trajectories to prioritize experience replay. 
\end{enumerate}

PPO+D is only in part on-policy as a fraction of its experience comes from a replay buffer and therefore was collected by an older version of the policy.
The importance sampling is limited to the action loss in PPO and does not adjust the target value in the value loss as in \citet{espeholt2018impala}.

We found that this new algorithm is capable of solving problems that are not solvable using normal PPO, behavioral cloning, GAIL, nor combining behavioral cloning and PPO. PPO+D is very easy to implement by only slightly modifying the PPO algorithm. Crucially, the learned policy is significantly different and  more efficient than the demonstration used in the training.

To test this new algorithm we created a benchmark of hard-exploration problems of varying levels of difficulty using the Animal-AI Olympics challenge environment \citep{beyret2019animal, crosby2019animal}.
All the tasks considered have random initial position and the PPO policy uses entropy regularization so that memorizing the sequence of actions of the demonstration will not suffice to complete any of the tasks. We also test PPO+D on a modified sparse version of the ReacherPyBulletEnv-v0 and LunarLander-v2 tasks based on the implementations available in \citet{brockman2016openai, coumans2017pybullet}.

\section{Related work}
Different attempts have been made to use demonstrations efficiently in hard-exploration problems.
In \citet{salimans2018learning} the agent is able to learn a robust policy only using one demonstration. The demonstration is replayed for $n$ steps after which the agent is left to learn on its own. By incrementally decreasing the number of steps $n$, the agent learns a policy that is robust to randomness (introduced in this case by using sticky actions or no-ops \citep{machado2018revisiting}, since the game is fundamentally a deterministic one). However, this approach only works in a fully deterministic environment since replaying the actions has the role of resetting the environment to a particular configuration. This method of resetting is obviously not feasible in a stochastic environment. 

 \citet{ecoffet2019go} introduces another algorithm that largely exploits the determinism of the environment by resetting to previously reached states.
It works by maximizing the diversity of the states reached. It is able to identify such diversity among states by down-sampling the observations, and by considering as different states only those observations that have a different down-sampled image. This works remarkably well in two dimensional environments, but is unlikely to work in three-dimensional, stochastic environments.

Self-supervised prediction approaches, such as \citet{pathak2017curiosity, burda2018exploration, schmidhuber2010formal, badia2020never}, have been used successfully in stochastic environments, although they are less effective in three-dimensional environments.
Another class of algorithms designed to solve exploration problems are count-based methods \citep{tang2017exploration}. These algorithms keep track of the states where the agent has been before (if we are dealing with a prohibitive number of states, the dimensions along which the agent moves can be discretized), and give the agent an incentive (in the form of a bonus reward) for visiting new states. This approach assumes we have a reliable way to track the position of the agent. This assumption is removed in the case of pseudo-counts when using neural density models \citep{bellemare2016unifying,
ostrovski2017count}. \citet{choi2018contingency} proposes an alternative method for localizing the agent based on an attention mask over pixels in the image used in predicting the action the agent took. Like \citet{pathak2017curiosity} we think this solution too is probably biased towards two dimensional environments.

An empirical comparison between these two classes of exploration algorithms was made in  \cite{baker2019emergent}, where agents compete against each other leveraging the use of tools that they learn to manipulate. They found the count-based approach works better when applied not only to the agent position, but also to relevant entities in the environment (such as the positions of objects). When only given access to the agent position, the Random Network Distillation  algorithm \citep{burda2018exploration} was found to lead to a higher performance.

Some other works focus on leveraging the use of expert demonstrations while still maximizing the reward. These methods allow the agent to outperform the expert demonstration in many problems.
\citet{hester2018deep} combines temporal difference updates in the Deep Q-Network (DQN) algorithm  with supervised classification of the demonstrator’s actions.
\citet{kang2018policy} proposes to effectively leverage available demonstrations to guide exploration through enforcing occupancy measure matching between the learned policy and current demonstrations.

Other approaches, such as \citet{duan2017one, zhou2019watch} pursue a meta-learning strategy where the agent learns to learn from demonstrations, such approaches are perhaps the most promising, but they require at least a demonstration for each task for a subset of all tasks. 

Generative adversarial imitation learning (GAIL) \citep{ho2016generative} has never been
successfully applied to complex partially observable environments that require memory \citep{paine2019making}. InfoGAIL \citep{li2017infogail} has been used to learn from images but, unlike in this work the policy does not require recurrence to complete the tasks.
Behavioral Cloning (BC) is the most basic 
imitation learning technique, it is equivalent to supervised classification of the demonstrator’s actions
\citep{rahmatizadeh2018vision}.
Both GAIL and BC have been used in the Obstacle Tower Challenge  \citep{juliani2019obstacle}, but are alone insufficient for solving hard-exploration problems \citep{obstacletowerunix}.

\citet{oh2018self} present an off-policy algorithm that learns to reproduce the agent's past good decisions. Their work focuses mainly on advantage actor-critic, but the idea was tested also with PPO.
 Instead, PPO+D utilizes importance sampling, leverages expert (human) demonstrations, uses prioritized experience replay based on the maximum value of each trajectory to overcome catastrophic forgetting and the ratio of demonstrations replayed during training can be explicitly  controlled.


\section{Demonstration guided exploration with PPO+D }

Our approach attempts to exploit the idea of combining demonstrations with the agent's own experience in an on-policy algorithm such as PPO. This approach is particularly effective for hard-exploration problems. One can view the replay of demonstrations as a possible trajectory of the agent in the current environment. This means that the only point we are interfering with the classical PPO is when sampling, which is substituted by simply replaying the demonstrations. 
A crucial difference between PPO+D and R2D3 \citep{paine2019making} is that we do consider sequences that contain entire episodes in them, and therefore using recurrence is much more straightforward. From the perspective of the agent it is as if it is always lucky when sampling the actions, and in doing so it is skipping the exploration phase. The importance sampling formula provides the average rewards over policy $\pi_{\theta'}$ given trajectories generated from a different policy $\pi_\theta(a|s)$:
\begin{equation}
    \mathbb{E}_{\pi_{\theta'}}[r_t]=\mathbb{E}_{\pi_\theta}\Big[ \frac{\pi_{\theta'}(a_t|s_t)}{\pi_{\theta}(a_t|s_t)} r_t \Big],
\end{equation}
where $r_t \sim R(r|a_t,s_t)$ is the environment reward given state $s_t$ and action $a_t$ at time $t$. $\mathbb{E}_{\pi_{\theta}}$ indicates that the average is over trajectories drawn from the parameterized policy $\pi_{\theta}$. The importance sampling term $\frac{\pi_{\theta'}(a_t|s_t)}{\pi_{\theta}(a_t|s_t)}$ accounts for the correction in the change of the distribution over actions in the  policy $\pi_{\theta'}(a_t|s_t)$. By maximizing $\mathbb{E}_{\pi_\theta}\Big[ \frac{\pi_{\theta'}(a_t|s_t)}{\pi_{\theta}(a_t|s_t)} r_t \Big]$ over the parameters $\theta'$ a new policy is obtained that is on average better than the old one. 
The PPO loss function is then defined as

\begin{equation}
\begin{gathered}
L_t^{PPO}(\theta) = \\ \mathbb{E}_t \Big[L_t^{CLIP}(\theta) + c_1 L_t^{VF}(\theta) + c_2 S^{\pi_{\theta}}(s_t, a_t)\Big],
\end{gathered}
\end{equation}

where 
\begin{equation}
\begin{gathered}
L_{t}^{CLIP}(s_t,a_t,\theta',\theta) = \\ \min\left(
\frac{\pi_{\theta'}(a_t|s_t)}{\pi_{\theta}(a_t|s_t)}  A^{\pi_{\theta}}(s_t,a_t), \;\;
g(\epsilon, A^{\pi_{\theta}}(s_t,a_t))
\right)
\end{gathered}
\end{equation}

and $c_1$ and $c_2$ are coefficients, $L_t^{VF}$ is the squared-error loss $(V_{\theta}(s_t)- V_{t}^{targ})^{2}$, $A^{\pi_{\theta}}$ is an estimate of the advantage function and $S$ is the entropy of the policy distribution. The entropy term is designed to help keep the search alive by preventing convergence to a single choice of
output, especially when several choices all lead to roughly the same reward \citep{williams1991function}.

Let $\mathcal{D}$ be the set of trajectories $\tau = (s_0, a_0, s_1, a_1, ...)$ for which we have a demonstration, then  $\pi_{\mathcal D}(a_t|s_t)=1$ for $(a_t,s_t)$ in $\mathcal{D}$ and $0$ otherwise.
This is a policy that only maps observations coming from the demonstration buffer to a distribution over actions.
Such distribution assigns probability one to the action taken in the demonstration and zero to all the remaining actions.
The algorithm decides where to sample trajectories from each time an episode is completed (for running out of time or for completing the task).
We define $ \mathcal D $ to be the set of all trajectories that can get replayed at any given time
$ \mathcal D = \mathcal D_{V} \cup \mathcal D_{R} $, where $\mathcal D_{V}$ are the trajectories collected prioritizing the value estimate ($V$ stands for value), and $\mathcal D_{R}$ ($R$ stands for reward) contains the initial human demonstration and successful trajectories the agent collects.
Since the sparse environments we tested PPO+D on only have one source of reward, any trajectory can either be considered successful in completing the task or unsuccessful.
The agent samples from the trajectories in 
$ \mathcal{D_R}$ with probability $\rho$ , $\mathcal{D_V}$ with probability $\phi$  and from the real environment  $ \text{Env} $ with probability $1- \rho -\phi$ subject to $\rho +\phi \leq 1$ .


The behavior of the policy can be defined as follows:\\
\begin{center}
$ \pi^{\phi,\rho}_{\theta}= 
\begin{cases}
     \pi_{\mathcal{D_R}},          & \text{if sampled from }  \mathcal{D_R}\\
     \pi_{\mathcal{D_V}},          & \text{if sampled from }  \mathcal{D_V}\\
    \pi_{\theta},              & \text{if sampled from Env }  
\end{cases}$
\end{center}

In PPO+D we substitute the current policy $\pi_{\theta}$ with the  policy $\pi^{\phi,\rho}_{\theta}$, since this is the policy used to collect the demonstration trajectories. The clipping term is then changed to correct for it,
\begin{equation}
\begin{gathered}
L_{t}^{PPO+D}(s_t,a_t,\theta',\theta) = \\ \min\left(
\frac{\pi_{\theta'}(a_t|s_t)}{\pi^{\phi,\rho}_\theta(a_t|s_t)}  A^{\pi^{\phi,\rho}_\theta}(s_t,a_t), \;\;
g(\epsilon, A^{\pi^{\phi,\rho}_\theta}(s_t,a_t))
\right).    
\end{gathered}
\end{equation}

\subsection{Self-imitation and prioritized experience replay}

We hold the total size of the buffers $|\mathcal D|$ constant throughout the training. At the beginning of the training we only provide the agent with one demonstration trajectory, so $| \mathcal D_{R} | = 1$ and $| \mathcal D_{V} | = | \mathcal D | - 1$ as
$|\mathcal D| =  |\mathcal D_{V}| + | \mathcal D_{R} |. $
$\mathcal D_{V}$ is only needed at the beginning to collect the first successful trajectories. As $|\mathcal D_R|$  increases and we have more variety of trajectories in $\mathcal D_R$ , we decrease the probability $\phi$ of replaying trajectories from $\mathcal D_V$. After  $|\mathcal D_R|$ is large enough, replaying the trajectories from  $\mathcal D_V$ is more of an hindrance. The main reason for treating these two buffers separately is to slowly anneal from one to the other avoiding the hindrance in the second phase.

$|\mathcal D_{V}|$ and $|\mathcal D_{R}|$ are the quantities that are annealed according to the following formulas, each time a new successful trajectory is found and is added  to $\mathcal D_{R}$ :

\begin{equation}\label{schedule}
\begin{gathered}
\rho = \rho + \frac{\phi_0}{| \mathcal D_{V} |_0}; \quad 
\phi = \phi - \frac{\phi_0}{| \mathcal D_{V} |_0}; \quad \\
|\mathcal D_{V}| = |\mathcal D_{V}| - 1; \quad  
|\mathcal D_{R}| = |\mathcal D_{R}| + 1 \quad
\end{gathered}
\end{equation}


where $| \mathcal D_{R} |_0$ and $| \mathcal D_{V} |_0$ are the maximum size respectively for $\mathcal D_{R}$ and $\mathcal D_{V}$ and $\phi_0$ is the value of $\phi$ at the beginning of the training. We define the probability of sampling trajectory $ \tau_i$ as $P(i) = \frac{
p_i^{\alpha}}{\sum_{k} p_k^{\alpha}}$, where $p_i= \max_t  V_{\theta}(s_t)$, and $\alpha$ is a hyperparameter. We shift the value of $p_i$ for all the trajectories in $\mathcal D_{V}$ so as to guarantee $p_i \geq 0$.  We only keep up to $|\mathcal D_{V}|$ unsuccessful trajectories at any given time.

\begin{figure}
    \centering
    \label{fig:my_label}
    \centerline{\scalebox{0.215}{\includegraphics{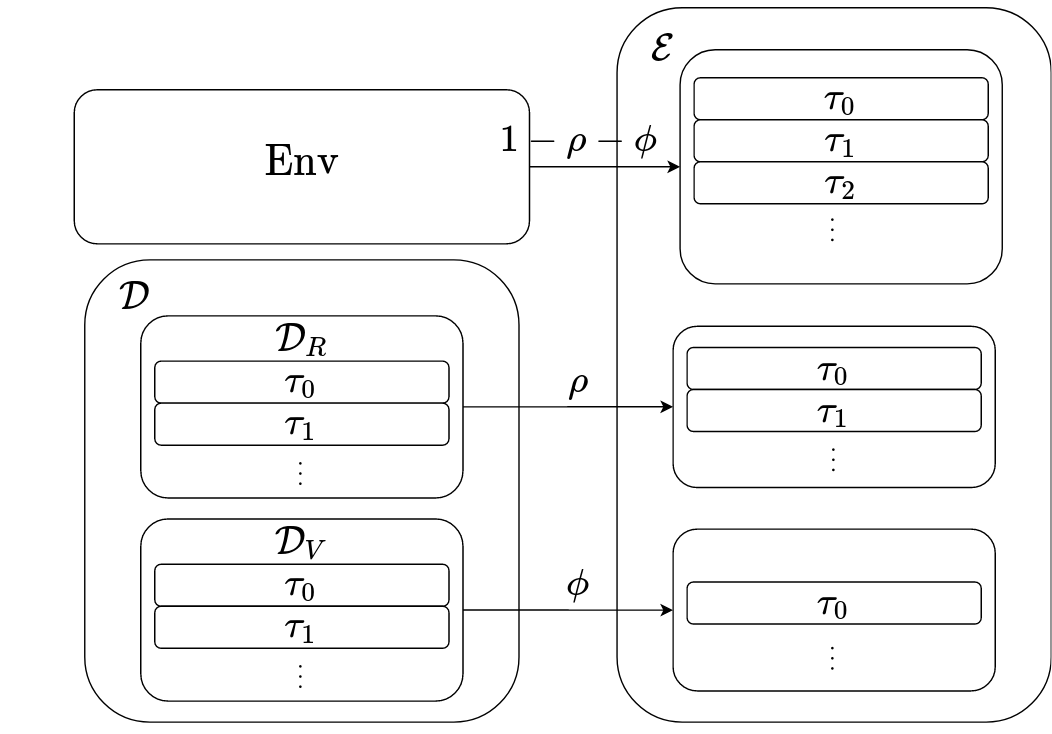}}}
  \caption{\textbf{Learner diagram.} The PPO+D learner samples batches that are a mixture of demonstrations and the experiences the agent collects by interacting with the environment. }
\end{figure}
  


Values are only updated for the replayed transitions.
Successful trajectories are similarly sampled from $\mathcal D_{R}$ with a uniform distribution (a better strategy could be to sample according to their length), and the buffer is updated following a FIFO strategy.
We introduced the value-based experience replay because we get to a level of complexity in some tasks  that we could not  solve by using self-imitation solely  based on the reward. These are the tasks that the agent has trouble solving even once because the sequence of actions is too long or complicated. We prefer the value estimate as a selection criteria rather than the more commonly used TD-error as we speculate it is more effective at retaining promising trajectories over long periods of time.
Crucially for the unsuccessful trajectories it is possible for the maximum value estimate not to be zero when the observations are similar to the ones seen in the demonstration.

\begin{algorithm}[H]
	\caption{PPO+D }
	\begin{algorithmic}
	
	    \STATE Initialize parameters $\theta$
	    \STATE Initialize replay buffer $ \mathcal D_V \leftarrow  \{\}  $
	    \STATE Initialize replay buffer $ \mathcal D_R \leftarrow  \{\tau_{\mathcal D} \}$
	    \STATE Initialize rollout storage $ \mathcal E \leftarrow  \{\}  $
	    \FOR {every update}
	      
	    \FOR {actors $1,2,\dots,N$}
	    \STATE Sample $\tau$ from $ \{\mathcal D_V,\mathcal D_R, \text{Env} \}$ based on $\rho$ and $\phi$
	    \IF{$\tau \in \mathcal D_R $}
	    \FOR {steps $1,2,\dots,T$}
		\STATE \begin{varwidth}[t]{\linewidth} Replay a transition from \par buffer $s_t,a_t, r_t, s_{t+1} \sim \pi_{\mathcal D_R}(a_t|s_t) $
		\end{varwidth}
		\STATE Store transition $\mathcal E \leftarrow \mathcal E \cup \{(s_t,a_t,r_t)\}$
		\ENDFOR
		\ELSIF{$\tau \in \mathcal D_V $}
	    \FOR {steps $1,2,\dots,T$}
		\STATE \begin{varwidth}[t]{\linewidth} Replay a transition from buffer \par $s_t,a_t, r_t, s_{t+1} \sim \pi_{\mathcal D_V}(a_t|s_t) $ 
		\end{varwidth}
		\STATE Store transition $\mathcal E \leftarrow \mathcal E \cup \{(s_t,a_t,r_t)\}$
		\ENDFOR
		\ELSIF {$\tau \in   \text{Env}$}
		\FOR {steps $1,2,\dots,T$}
		\STATE \begin{varwidth}[t]{\linewidth} Execute an action in the environment \par $s_t,a_t, r_t, s_{t+1} \sim \pi_{\theta}(a_t|s_t) $
		\end{varwidth}
		\STATE Store transition $\mathcal E \leftarrow \mathcal E \cup \{(s_t,a_t,r_t)\}$
		\ENDFOR
		\ENDIF
		\STATE Compute advantages estimates $\hat{A}_1,\hat{A}_2, \dots, \hat{A}_T$ 
		\ENDFOR
		
		\STATE \begin{varwidth}[t]{\linewidth} Optimize $L^{PPO}$ wrt $\theta $, with $K$ epochs and \par minibatches size $M \leq NT$
		\end{varwidth}
		\STATE $\theta \leftarrow \theta - \eta \nabla_{\theta} L^{PPO}$
        \STATE Empty rollout storage  $ \mathcal E \leftarrow  \{\}  $
        \STATE Update $|\mathcal D_{V}|$, $|\mathcal D_{R}|$, $\rho$ and $\phi$ according to Equations \ref{schedule}
		\ENDFOR
\end{algorithmic} 
\end{algorithm} 

 We think that this plays a role in countering the effects of catastrophic forgetting, thus allowing the agent to combine separately learned sub-behaviors in a successful policy. We illustrate this with some example trajectories of the agent shown in the Appendix in Figure \ref{fig:S1}.
The value estimate is noisy and as a consequence of that, trajectories that have a low maximum value estimate on first visits may not be replayed for a long time or never as pointed out in  \cite{schaul2015prioritized}. However, for our strategy to work it is enough for some of the promising trajectories to be collected and replayed by this mechanism.

\section{Experiments}

\subsection{The Animal-AI Olympics challenge environment}
The recent successes of deep reinforcement learning (DRL)  \citep{mnih2015human, silver2017mastering, schulman2017proximal,schrittwieser2019mastering,srinivas2020curl} have shown the potential of this field, but at the same time have revealed the inadequacies of using games (such as the ATARI games \citep{bellemare2013arcade}) as a benchmark for intelligence. These inadequacies have motivated the design of more complex environments that will provide a better measure of intelligence.

 The Obstacle Tower Environment \citep{juliani2019obstacle},
the Animal AI Olympics \citep{crosby2019animal},
the Hard-Eight Task Suite \citep{paine2019making} and the DeepMind Lab \citep{beattie2016deepmind} all exhibit sparse rewards, partial observability and highly variable initial conditions.
For some tests we use the Animal-AI Olympics challenge environment.
The aim of the Animal-AI Olympics is to translate animal cognition into a benchmark of cognitive AI \citep{crosby2019animal}.

The environment contains an agent enclosed in a fixed size arena.  Objects can spawn in this arena, including positive and negative rewards (green, yellow and red spheres) that the agent must obtain or avoid. This environment has basic physics rules and a set of objects such 
as food, walls, negative-reward zones, 
movable blocks and more. The playground 
can be configured by the participants and 
they can spawn any combination of objects 
in preset or random positions. We take advantage of the great flexibility allowed by this environment to design hard-exploration problems for our experiments. 

We also included some experiments on tasks that have already been extensively studied in the literature: the ReacherPyBulletEnv-v0 \cite{coumans2017pybullet} and LunarLander-v2 \cite{brockman2016openai} tasks. However, because the use of PPO+D is only justified when dealing with very sparse reward environments we modified the reward function of these tasks to fulfill this criteria.

\begin{figure*}
    \centering
    \centerline{\scalebox{0.4}{\includegraphics{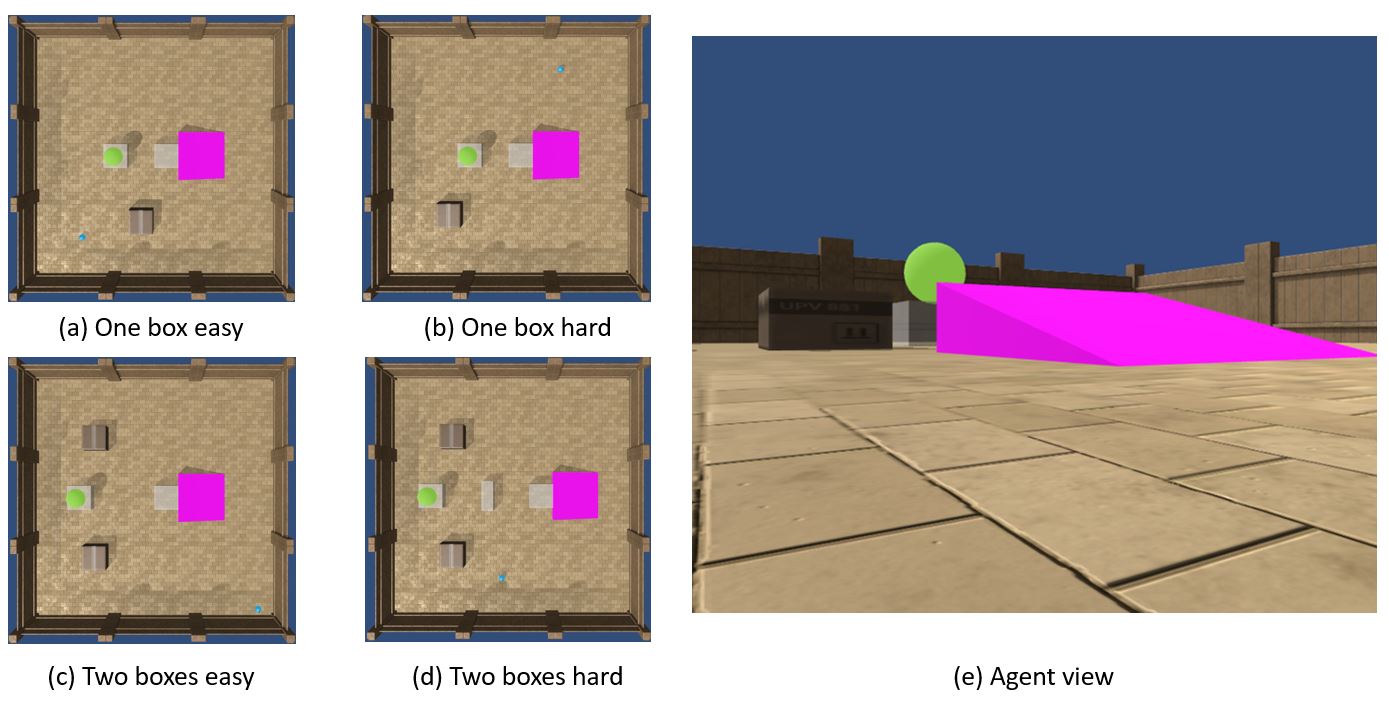}}}
    \caption{ \textbf{Tasks.} 
In each of the tasks there is only one source of reward and the position of some of the objects is random, so each episode is different. The agent has no access to the aerial view, instead it partially observes the world through a first person view of the environment. All of the tasks are either inspired or adapted from the test set in the Animal-AI Olympics competition. A video showing the experimental results is available at \url{https://www.youtube.com/playlist?list=PLBeSdcnDP2WFQWLBrLGSkwtitneOelcm-} .}
\label{fig:maps}
\end{figure*}

\subsection{Experimental setting}

We designed our experiments in order to answer the following questions:
 Can the agent learn a policy in a non-deterministic hard-exploration environment only with one  human demonstration?
 Is the agent able to use the human demonstration to learn to solve a problem with different initial conditions (agent and boxes initial positions) than the demonstration trajectory?
 How does the hyperparameter $\phi$ influence the performance during training?
The four tasks in the Animal AI Olympics environments we used to evaluate the agent are described as follows:
\setlist{nolistsep}
\begin{itemize}[noitemsep]

\item
\textbf{One box easy}

The agent has to move a single box, always spawned at the same position, in order to bridge a gap and be able to access the reward once it climbs up the ramp (visible in pink). The agent can be spawned in the range $(X:0.5-39.5,Y:0.5-39.5 )$ if an object is not already present at the same location (Fig. \ref{fig:maps}a).

\item
\textbf{One box hard}

The agent has to move a single box in order to bridge a gap and be able to access the reward, this time two boxes are spawned at any of four positions $A:(X:10, Y:10), B:(X:10, Y:30), C:(X:30, Y:10), D:(X:30, Y:30)$. The agent can be spawned in the range $(X:0.5-39.5,Y:0.5-39.5 )$ if an object is not already present at the same location (Fig. \ref{fig:maps}b).

\item
\textbf{Two boxes easy}

The agent has to move two boxes in order to bridge a larger gap and be able to access the reward, this time two boxes are spawned at any of four positions $A:(X:10, Y:10), B:(X:10, Y:30), C:(X:30, Y:10), D:(X:30, Y:30)$. The agent can be spawned in the range $(X:15.0-20.0 ,Y:0.5-15.0 )$ if an object is not already present at the same location (Fig. \ref{fig:maps}c).

\item
\textbf{Two boxes hard }

The agent has to move two boxes in order to bridge a larger gap and be able to access the reward. Two boxes are spawned at two fixed positions $A:(X:10, Y:10), B:(X:10, Y:30)$. A wall acts as a barrier in the middle of the gap, to prevent the agent from "surfing" a single box. The agent can be spawned in the range $(X:15.0-20.0 ,Y:5.0-10.0 )$ if an object is not already present at the same location (Fig. \ref{fig:maps}d).
\end{itemize}

For the Sparse LunarLander-v2 tasks the reward has zero value if the agent is unable to land and one otherwise.
For the ReacherPyBulletEnv-v0 task the common reward function contains a term proportional to the negative euclidean distance between the tip of the robot arm and the target position. 
In the Sparse ReacherPyBulletEnv-v0 the reward is modified as follows:

\begin{center}
$r = 
\begin{cases}
     T - d,          & \text{if } d \leq T  \\
     0,          & \text{if }  d > T\\
    
\end{cases}$
\end{center}

with $T \geq 0 $ and $d \geq 0 $, where $d$ is the euclidean distance between tip of the arm and target position, and $T$ is a threshold value. For our experiments we chose $T = 1$.  


\section{Results}

\subsection{Comparison with baselines and generalization }

\begin{figure*}[h]
    \centerline{\scalebox{0.2}{\includegraphics{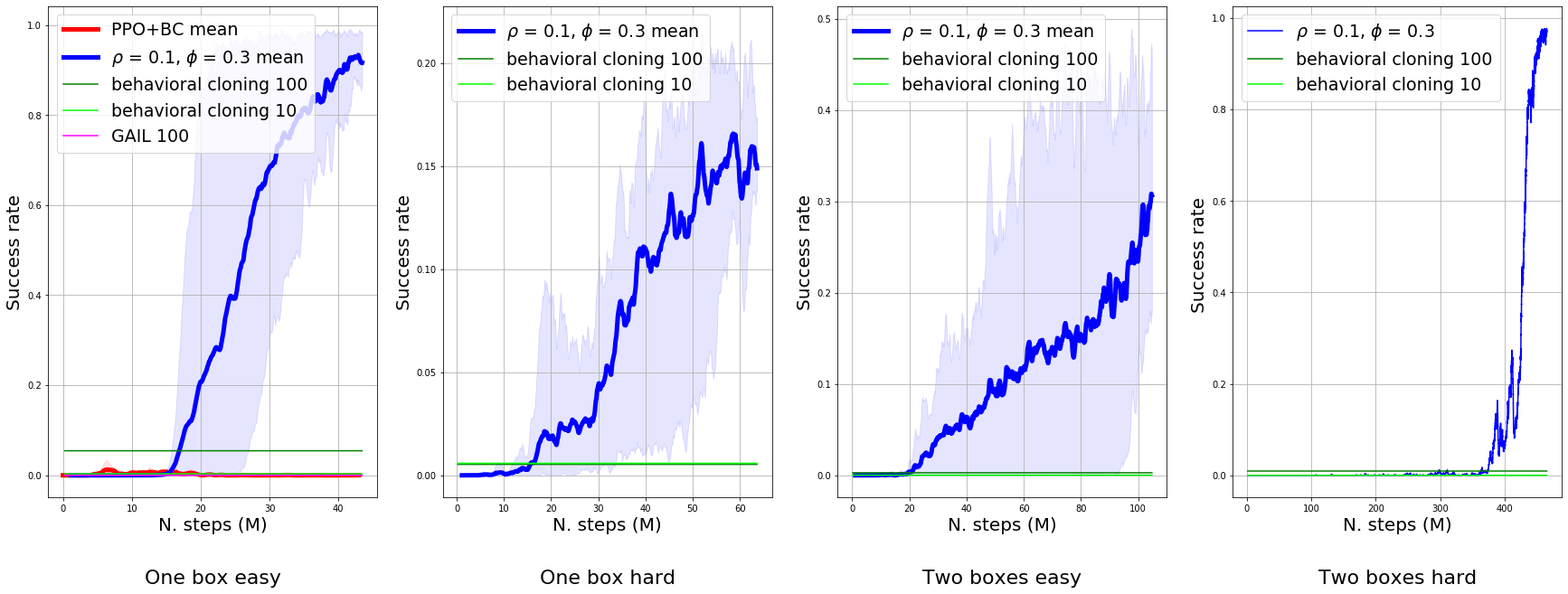}}}
    \caption{ \textbf{Experiments.} Performance of  behavioral cloning with ten and a hundred recorded human demonstrations and PPO+D with $\rho = 0.1, \phi = 0.3$ and just one demonstration. The curves represent the mean, min and max performance for each of the baselines across 5 different seeds, except for vanilla PPO, GAIL and PPO+BC (3 seeds) where the reward is consistently close to zero. The BC agent sporadically obtains some rewards. GAIL with a hundred demonstrations never achieves any reward. PPO+BC has only access to one demonstration, like PPO+D. It occasionally solves the task but it is unable to achieve high performance. }
    \label{fig:expe1}
   
\end{figure*}

In Figure \ref{fig:expe1} we compare PPO+D with parameters $\rho = 0.1, \phi = 0.3$ to the behavioral cloning baselines (with $100$ and $10$ demonstrations), to GAIL (with $100$ demonstrations) and with PPO+BC. PPO+BC combines PPO and BC in an a similar way to PPO+D: with probability $\rho$ a sample is drawn from the demonstrations and the policy is updated using the BC loss (see Section \ref{chap:hyperpara} in the Appendix for more details).  The value loss function remains unchanged during the BC update. 

We tested the GAIL implementation on a simple problem to validate it (see Section \ref{chap:gail_test} in the Appendix).  For behavioral cloning we trained for 3000 learner steps (updates of the policy) with learning rate $10^{-5}$. It is clear that PPO+D is able to outperform the baseline in all four problems. The performance of PPO+D varies considerably from task to task. This reflects the considerable difference in the range of the initial conditions for different tasks. In the tasks "One box easy" and "Two boxes hard" only the position of the agent sets different instances of the task apart. The initial positions of the boxes only play a role in the tasks "One box hard" and "Two boxes easy". Under closer inspection we could verify that in these two tasks the policy fails to generalize to configurations of boxes that are not seen in the demonstration, but does generalize well in all tasks for very different agent starting positions and orientations.

This could be because there is a smooth transition between some of the initial conditions. Due to this fact, if the agent is able to generalize even only between initial conditions that are close, it will be able to progressively learn to perform well for all initial conditions starting from one demonstration. In other words the agent is automatically using a curriculum learning strategy, starting from the initial conditions that are closer to the demonstration. This approach fails when there is an abrupt transition between initial conditions, such as for different boxes configurations. 

During training we realized that the policy exploited the 3D engine in the task "Two boxes easy" as it managed to "surf" one of the boxes, in this way avoiding to move the remaining box (a similar behavior was reported in \cite{baker2019emergent}).  To avoid this trick and make sure the agent is capable of such more complex behavior we introduced "Two boxes hard". We decided to reduce the range of the initial conditions in this last task, as we already verified that the agent can learn from such variable conditions in tasks "One box hard" and "Two boxes easy". This last experiment only tests the agent for a more complex behavior.
In the tasks "One box hard" and "Two boxes easy" the agent could achieve higher performance given more training.

\begin{figure*}[h]
    \centerline{\scalebox{0.2}{\includegraphics{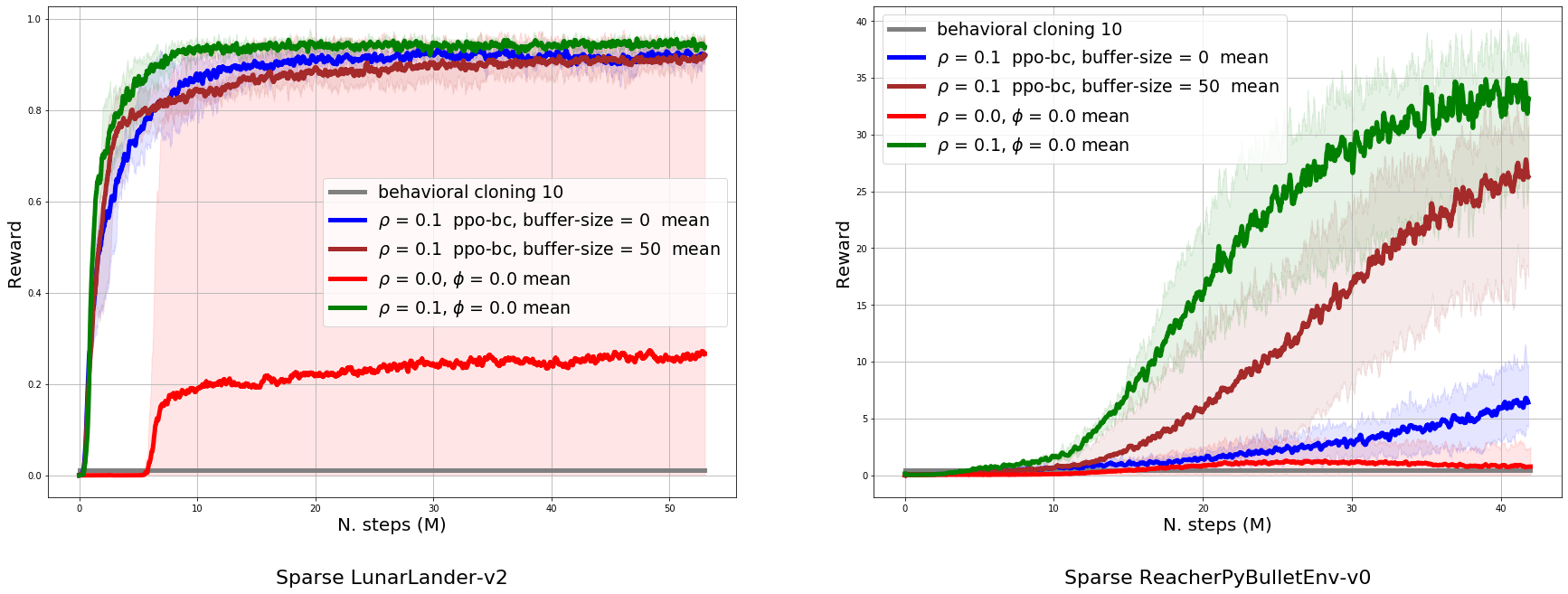}}}
    \caption{ \textbf{Experiments.} Performance of  PPO+D (green), PPO+BC with reward buffer (buffer size =50, brown) and without it (buffer size =0, blue) vanilla PPO (red) and normal behavior cloning (gray). For Sparse ReacherPyBulletEnv-v0 we chose the hyperparameters $\rho =0.1,  \phi = 0.0$ and for Sparse LunarLander-v2 $\rho =0.3,  \phi = 0.0$, which we found to be optimal for both PPO+D and PPO+BC. For both experiments the agent has ten human demonstrations available to learn from. The curves show the mean min and max reward for each of the baselines across 5 different seeds.}
    \label{fig:expe3}
   
\end{figure*}

\begin{figure*}[h]

    \centerline{\scalebox{0.2}{\includegraphics{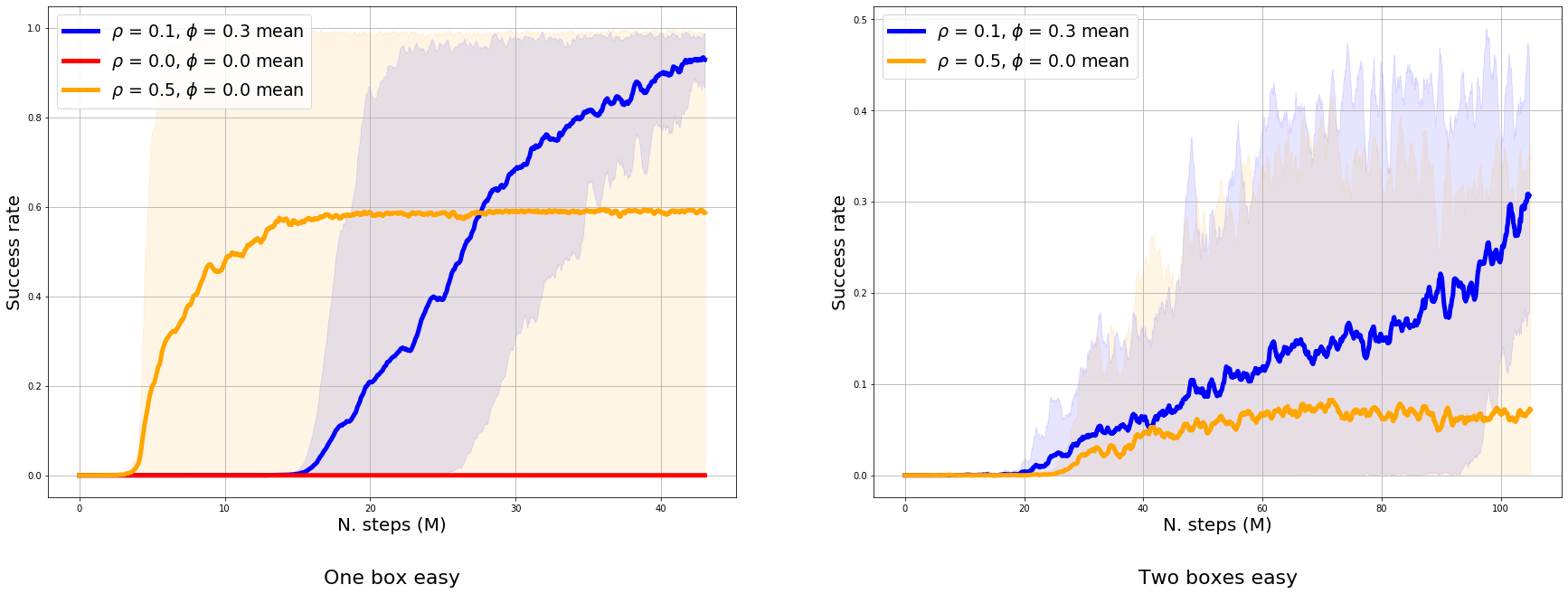}}}
   \caption{ \textbf{Experiments.} Performance for vanilla PPO ($\rho = 0.0, \phi = 0.0$), PPO+D with $\rho = 0.5, \phi = 0.0$ and PPO+D with $\rho = 0.1, \phi = 0.3$ on the tasks "One box easy" and "Two boxes easy" using a single demonstration.
   Some of the curves overlap each other as they receive zero or close to zero reward. Vanilla PPO never solves the task. The curves show the mean, min and max performance across 5 seeds.}
   \label{fig:expe2}

  \end{figure*}
  
In Figure \ref{fig:expe3} we compare PPO+D with PPO+BC and vanilla PPO on the tasks Sparse LunarLander-v2 and Sparse ReacherPyBulletEnv-v0. These  tasks are considerably easier than the Animal AI Olympics tasks but are useful to underlie the difference in performance between PPO+D and PPO+BC. 
For these experiments we used ten demonstrations.
For both tasks we first tuned the $\rho$ hyperparameter for both PPO+D and PPO+BC. For PPO+BC we consider two versions, one in which it keeps adding successful trajectory to a replay buffer in the same way PPO+D does, in the other it does not have such replay buffer.
In these two tasks we do not employ the value buffer as this is of any use only in very hard exploration tasks.
Surprisingly even vanilla PPO is able to solve Sparse LunarLander-v2 sporadically (in one seed over five), although its performance exhibits high variance.
Both variants of PPO+BC successfully learn the task, the difference with PPO+D is small but consistent over all seeds. In the more challenging task Sparse ReacherPyBulletEnv-v0 the differences in performance between PPO+D and PPO+BC with and without reward buffer are considerably amplified.

We emphasise that PPO+D is designed to perform well on hard-exploration problems with stochastic environment and variable different conditions, we tested it on the  of the Atari environment BreakoutNoFrameskip-v4 and conclude that it does not lead to better performance than vanilla PPO when the reward is dense. PPO+D also learns to complete the task although more slowly than PPO.

\subsection{The role of \texorpdfstring{$\phi$} \\  and    the value-buffer  \texorpdfstring{$\mathcal{D_V}$} \\ }
\label{chap:comparison}
In Figure \ref{fig:expe2} we mainly compare PPO+D with $\rho = 0.1, \phi = 0.3$ to  $\rho = 0.5, \phi = 0.0$. Crucially, the second parameter configuration does not use the buffer $\mathcal D_V$. It is evident in the task "Two boxes easy" that $\mathcal D_V$ is essential for solving harder exploration problems. In the "One box easy" task we can see that occasionally on easier tasks not having $\mathcal D_V$ can result in faster learning. However, this comes with a greater variance in the training performance of the agent, as sometimes it completely fails to learn.

In Figure \ref{fig:expe2} we also run vanilla PPO $(\rho = 0.0, \phi = 0.0)$ on the task "One box easy" and establish its inability to solve the task even once on any seed and any initial condition. This being the easiest of all tasks, we consider unlikely the event of vanilla PPO successfully completing any of the other harder tasks. We defer an ablation study of the parameter $\rho$ to section \ref{chap:ablation} in the Appendix. Although we can not make a direct comparison with \cite{paine2019making}, we think it is useful to underline some of the differences between the two approaches both in the problem setting and in the performance.
We attempted to recreate the same complexity of the tasks on which \cite{paine2019making} was tested. In the article, variability is introduced in the tasks on multiple dimensions such as position and orientation of the agent, shape and color of objects, colors of floor and walls, number of objects in the  environment and position of such objects. The range of the initial conditions for each of these factors was not reported. In our work we  change the initial position and orientation of the agent as well as the initial positions of the boxes. As for the number of demonstrations in \cite{paine2019making} the agent has access to a hundred demonstrations, compared to only one in our work. In the present article, the training time ranges from under 5 millions frames to 500 millions, whereas in \cite{paine2019making} it ranges from 5 billions to 30.
Although we adopted the use of the parameter $\rho$ from \cite{paine2019making} its value differs considerably, which we attribute to the difference between the two algorithms: one being based on PPO, the other on DQN.

It is worth pointing out that PPO+D performance is likely to suffer in environments where the reward function is also stochastic (e.g. imperfect-information games). This is due to the fact that the importance sampling in the current architecture can not compensate for the difference between the training data distribution, which is biased by the replay, and the real environment data distribution.   


\section{Conclusion}

We introduce PPO+D, an algorithm that uses a single demonstration to explore more efficiently in hard-exploration problems. We further improve on the algorithm by introducing two replay buffers: one containing the agent own successful trajectories as it collects these over the training and the second collecting unsuccessful trajectories with a high maximum value estimate. In the second buffer the replay of the trajectories is prioritized  according to the maximum estimated value. We show that training with both these buffers solves, to some degree, all tasks the agent was presented with. We also show that vanilla PPO as well as PPO+D without the value-buffer fails to learn the same tasks. In the article, we justify the choice of such adjustments as measures to counter catastrophic forgetting, a problem that afflicts PPO particularly. The present algorithm suffers some limitations as currently it fails to generalize to all variations of some of the problems, yet it achieves to solve several very hard exploration problems with a single demonstration. The source code is available at \url{https://github.com/compsciencelab/ppo_D}.



\bibliographystyle{icml2021}
\bibliography{main.bib}

\clearpage

\appendix
\section{Training details} 

\label{chap:hyperpara}

 For the training we used 14 parallel environments and we compute the gradients using the Adam optimizer \citep{kingma2014adam} with fixed learning rate of $10^{-5}$. In the Animal AI Olympics environment the agent perceives the environment through a 84 by 84 RGB pixels observations in a stack of 4. At each time-step the agent is allowed to take one of nine actions.
We use the network architecture proposed in  \cite{pytorchrl} which includes a gated recurrent unit (GRU) \citep{gru} with a hidden layer of size 256.
 We ran the experiments on machines with 32 CPUs and 3 GPUs, model GeForce RTX 2080 Ti.
The experiments for the Animal-AI Olympics environment where carried out with the following hyperparameters.

\begin{table}[H]
\caption{Model and PPO Hyperparameters }
\centering
 \begin{tabular}{c c }
  \toprule
  Parameter & Value \\
 \midrule
  clip-param & 0.15 \\
  gamma & 0.998 \\ 
  frame-skip & 2 \\
  frame-stack & 4 \\
  num-steps & 1000 \\
  num-mini-batch & 6 \\
  entropy-coef & 0.02 \\
  value-loss-coef & 0.1 \\
  num-processes & 14 \\
  learning rate & 1e-5 \\
  eps  \text{(RMSprop optimizer epsilon)} & 1e-5 \\
  alpha \text{(RMSprop optimizer apha)} & 0.99 \\
  gae-lambda & 0.95 \\
  max-grad-norm & 0.5 \\
  ppo-epoch & 4 \\
  
\end{tabular}
\label{Tab:Tab_1}
\end{table}

Different hyperparameters were used for the experiments in the ReacherPyBulletEnv-v0 and LunarLander-v2 environments. Table \ref{Tab:Tab_2} shows the hyperparameters used for ReacherPyBulletEnv-v0. When training PPO+D, PPO+BC and vanilla PPO for LunarLander-v2, only slight changes were made in comparison to the ReacherPyBulletEnv-v0 hyperparameters: the number of processes (num-processes) changed from 64 to 32 and the entropy coefficient (entropy-coef) changed from 0.02 to 0.01. For both these tasks the policy is not recurrent as for the Animal AI Olympics tasks.

\begin{table}[H]
\caption{Model and PPO Hyperparameters (ReacherPyBulletEnv-v0)}
\centering
 \begin{tabular}{c c }
  \toprule
  Parameter & Value \\
 \midrule
  clip-param & 0.2 \\
  gamma & 0.99 \\ 
  frame-skip & 1 \\
  frame-stack & 1 \\
  num-steps & 2048 \\
  num-mini-batch & 32 \\
  entropy-coef & 0.02 \\
  value-loss-coef & 0.3 \\
  num-processes & 64 \\
  learning rate & 2e-4 \\
  eps  \text{(RMSprop optimizer epsilon)} & 1e-5 \\
  alpha \text{(RMSprop optimizer apha)} & 0.99 \\
  gae-lambda & 0.95 \\
  max-grad-norm & 0.5 \\
  ppo-epoch & 10 \\
  
\end{tabular}
\label{Tab:Tab_2}
\end{table}

When training PPO+BC the loss combines the PPO loss and the behavior cloning loss in the following way: 


\begin{center}
$L = 
\begin{cases}
     L_{PPO},          & \text{if } \tau_i \sim \mathcal D  \\
     L_{BC},          & \text{if }  \tau_i \sim \text{Env}\\
    
\end{cases}$
\end{center}

where
$L_{BC}$ is defined as the cross-entropy loss
\begin{equation*}
     L_{CE} = -\sum\limits_{t=1}^{n} a_t \log(\pi(a_t|s_t)) 
\end{equation*}
when the action space is discrete, and as the mean squared error:
\begin{equation*}
     L_{MSE} = \dfrac{1}{n}\sum\limits_{t=1}^{n}(\pi(a_t|s_t) - a_t)^2  \\
\end{equation*}
when the action space is continuous. Note that in the above equations $a_t$ refers to the action taken in the demonstration.

For the Animal AI Olympics tasks we performed no hyperparameter search over the replay ratios $\phi$ and $\rho$ but set them to a reasonable number. We found other configurations of these parameters to be sometimes more efficient in the training, such for example setting  $\rho = 0.5$ and $\phi = 0.0$ in the task "One box easy". The parameters we ran all the experiments with have been chosen because they allow to solve all of the experiments with one demonstration. We did run a hyperparameter search for the parameter $\rho$ for the LunarLander-v2 and ReacherPyBulletEnv-v0 task for both PPO+D and PPO+BC. The rest of the hyperparamters were adopted from \citet{schulman2017proximal}.

In computing the probability of a trajectory to be replayed $P(i) = \frac{
p_i^{\alpha}}{\sum_{k} p_k^{\alpha}}$, $\alpha = 10$.
The total buffer size is  $|\mathcal D| = 51$ with $|\mathcal D_V|_0 = 50$ plus the human generated trajectory. $|\mathcal D_R|_0 = 51$ meaning once the agent collects $50$ successful trajectories, new successful trajectories overwrite old ones, following a FIFO strategy and no trajectory is replayed from the value-buffer. The implementation used is  based on the repository 
\cite{pytorchrl}.
On our infrastructure we could train at approximately a speed of 1.3 millions frames per hour.  The code, pre-trained models, data-set of arenas used for training are available at \url{https://doi.org/10.6084/m9.figshare.13853039.v1}

\section{Hyperparameters ablation }
\label{chap:ablation}
In this section we present the results of four different experiments on a variation of the "One box easy task" where the agent position does not change across episodes and it is shared with the demonstration. We test on this variation of the task because it is one of the simplest problems we can use to test PPO+D performance. We only perform the ablation study on $\rho$ because $\phi$ is harder to test: it is indispensable for solving difficult tasks but it can slow down the performance on easy tasks. This being an easy task, the results obtained, do not provide any insights on the effect of $\phi$ in harder problems (as shown in Figure \ref{fig:expe2} ). The following figure shows the performance of the PPO+D algorithm where the $\rho$ parameter is changed and $\phi = 0$. Interestingly we observe that, among the values chosen, the performance peaks for $\rho = 0.3$. We hypothesize that lower $\rho$ values have worse performance because the interval between demo replays is so large that allows the network to forget the optimal policy learned with the demonstrations. On the other side, higher values of $\rho$ are even more counterproductive as they prevent the agent from learning from its own experience, most critically learning what not to do.    

Our implementation of GAIL based on \cite{li2017infogail} was trained with the following hyperparameters besides the PPO parameters in Table \ref{Tab:Tab_1}.

\section{GAIL test}
\label{chap:gail_test}
\begin{figure}[H]
    \centering
    \centerline{\scalebox{0.3}{\includegraphics{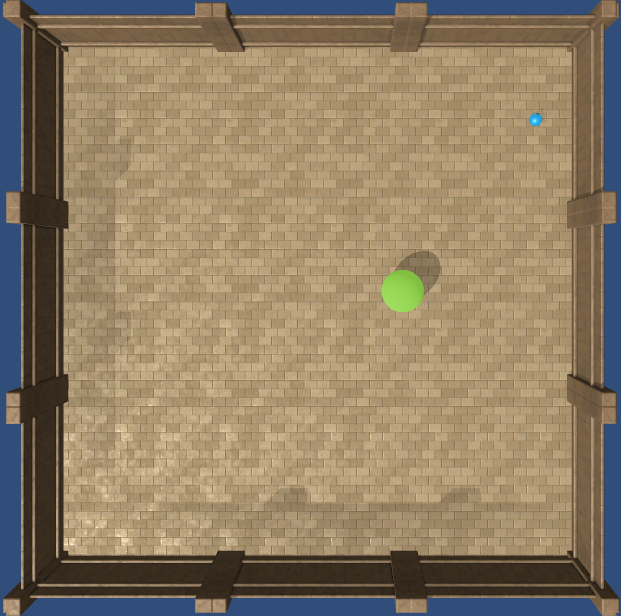}}}
    \caption{ \textbf{Food collection task.} 
In this task the the agent is spawned into the arena with one green ball. The green food size and position are set randomly at the beginning of each episode. The episode ends when the green food is collected. }
\label{fig:food_simple}
\end{figure}

To verify the correctness of our GAIL implementation we use for the experiments in Figure \ref{fig:expe2} we test it on a simple task in the Animal-AI environment. The task is shown in Figure \ref{fig:food_simple}, it consists in collecting green food of random size and position.

\begin{table}[h]
\caption{GAIL Hyperparameters}
\centering
 \begin{tabular}{c c }
  \toprule
  Parameter & Value \\
 \midrule
  scaling-factor & 0.001 \\
  gail-epoch & 0.4 \\ 
  gail-batch-size & 200 \\

\end{tabular}
\end{table}

\begin{figure}[H]
    \centering

    \centerline{\scalebox{0.3}{\includegraphics{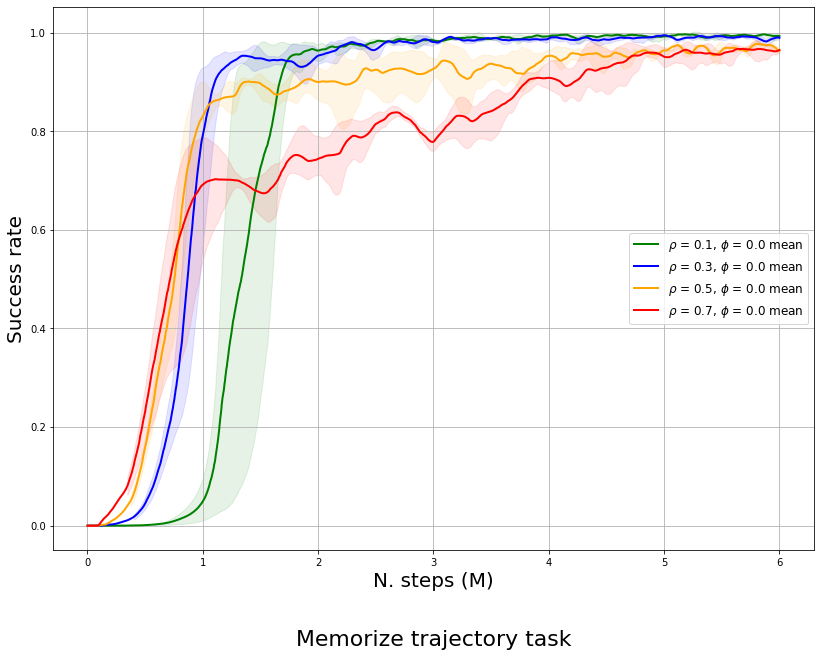}}}
   \caption{ \textbf{Ablation study} Performance for PPO+D with $\rho = 0.1, \phi = 0.0$, $\rho = 0.3, \phi = 0.0$, $\rho = 0.5, \phi = 0.0$ and $\rho = 0.5, \phi = 0.0$ and PPO+D with $\rho = 0.7, \phi = 0.0$, on a variation of the "One box easy" task were the initial position of the agent is fixed. The curves represent the mean, min and max performance for each of the baselines across 3 different seeds.}
   \label{fig:ablation}

  \end{figure}

\begin{table}[h]
\caption{Performance on the "Food collection task" }
\centering
 \begin{tabular}{c c c}
  \toprule
  Method & Avg. Success rate & Std. \\
 \midrule
  GAIL & 0.997 & 0.045 \\
  BC & 0.617 & 0.487 \\

\end{tabular}
\label{Tab:Tab_3}
\end{table}

\vfill\clearpage
\onecolumn
\section{Analysis of the effect of the value-buffer}

\begin{figure}[H]
    \centering

    \centerline{\scalebox{0.5}{\includegraphics{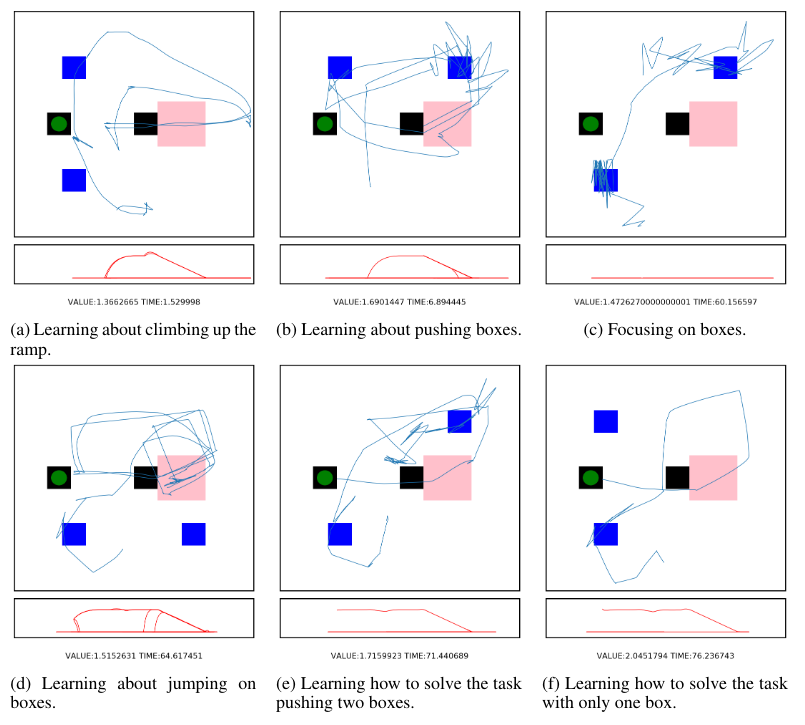}}}
    \caption{ \textbf{Sub-behaviors.} Trajectories the agent played on the task "Two boxes easy". In each of the figure the upper part shows the movements of the agent on the X-Y plane while the lower part shows the movement on the X-Z plane. The images are ordered by the time they were executed in the training in millions of frames. }
    \label{fig:S1}

  \end{figure}

The value-buffer experience replay creates an incremental curriculum for the agent to learn, keeping different trajectories that achieved an high maximum value in the buffer incentives the agent to combine these different sub-behaviors e.g. pushing the blocks and going up the ramp.

\end{document}